\documentclass{article}

\usepackage[nonatbib, preprint]{neurips_2020}

\usepackage[utf8]{inputenc} 
\usepackage[T1]{fontenc}    
\usepackage{hyperref}       
\usepackage{url}            
\usepackage{booktabs}       
\usepackage{amsfonts}       
\usepackage{nicefrac}       
\usepackage{microtype}      
\usepackage{graphicx}
\usepackage{color}
\usepackage{hyperref}

\graphicspath{ {./images} }

\title{Adaptive Compact Attention For Few-shot Video-to-video Translation}

\author{%
Risheng Huang$^{1,2}$\quad Li Shen$^2$\quad Xuan Wang$^2$\quad Cheng Lin$^1$\quad Hao-Zhi Huang$^2$ \\[0.3em]
$^1$The University of Hong Kong \quad\quad $^2$Tencent AI Lab\\
 \texttt{huangrs@connect.hku.hk, mathshen@tencent.com, cvxwang@tencent.com}\\
 \texttt{chlin@hku.hk, matthzhuang@tencent.com} 
}

\begin{document}

\maketitle

\begin{abstract}

This paper proposes an adaptive compact attention model for few-shot video-to-video translation\footnote{Video available: \url{https://youtu.be/1OCFbUrypKQ}}.  Existing works in this domain only use features from pixel-wise attention without considering the correlations among multiple reference images, which leads to heavy computation but limited performance. Therefore, we introduce a novel adaptive compact attention mechanism to efficiently extract contextual features jointly from multiple reference images, of which encoded view-dependent and motion-dependent information can significantly benefit the synthesis of realistic videos. Our core idea is to extract compact basis sets from all the reference images as higher-level representations. To further improve the reliability, in the inference phase, we also propose a novel method based on the Delaunay Triangulation algorithm to automatically select the resourceful references according to the input label. We extensively evaluate our method on a large-scale talking-head video dataset and a human dancing dataset; the experimental results show the superior performance of our method for producing photorealistic and temporally consistent videos, and considerable improvements over the state-of-the-art method.

\end{abstract}

\section{Introduction}\label{introduction-sec}

Video-to-video (vid2vid) translation, aiming to synthesize a photorealistic video guided by a corresponding semantic video, is one of the essential technologies for massive applications in computer vision, computer graphics, and movie industry. A common limitation of most vid2vid translation methods lies in the fact that it can only generate videos that are similar to the training data. To address this issue, Wang et al. \cite{DBLP:conf/nips/Wang0TLCK19} proposes a few-shot vid2vid translation method, of which the key idea is to dynamically generate the weights of a Generative Adversarial Network (GAN) \cite{DBLP:conf/nips/GoodfellowPMXWOCB14} using the features extracted by pixel-wise attention from reference images. To our best knowledge, this is the only work that models vid2vid translation in a few-shot manner, enabling the network to generate videos from unseen images. However, the proposed pixel-wise attention model does not explicitly consider the higher-level information encoded in multiple images, leading to heavy computation but limited performance.

We observe the context of reference images is highly informative for synthesizing realistic videos in few-shot vid2vid translation, since it captures the motion-dependent and view-dependent information. However, due to the non-local property, simply incorporating a self-attention model to capture the contextual information from multiple images is expensive, i.e. consuming a substantial amount of time and memory, which creates a bottleneck of the performance and applicability. Besides, the attention computation process in vid2vid is individual for each reference image. In other words, no coordination is conducted over different reference images to avoid extracting redundant information.

\begin{figure}[t]
    \centering
    \includegraphics[width = \linewidth]{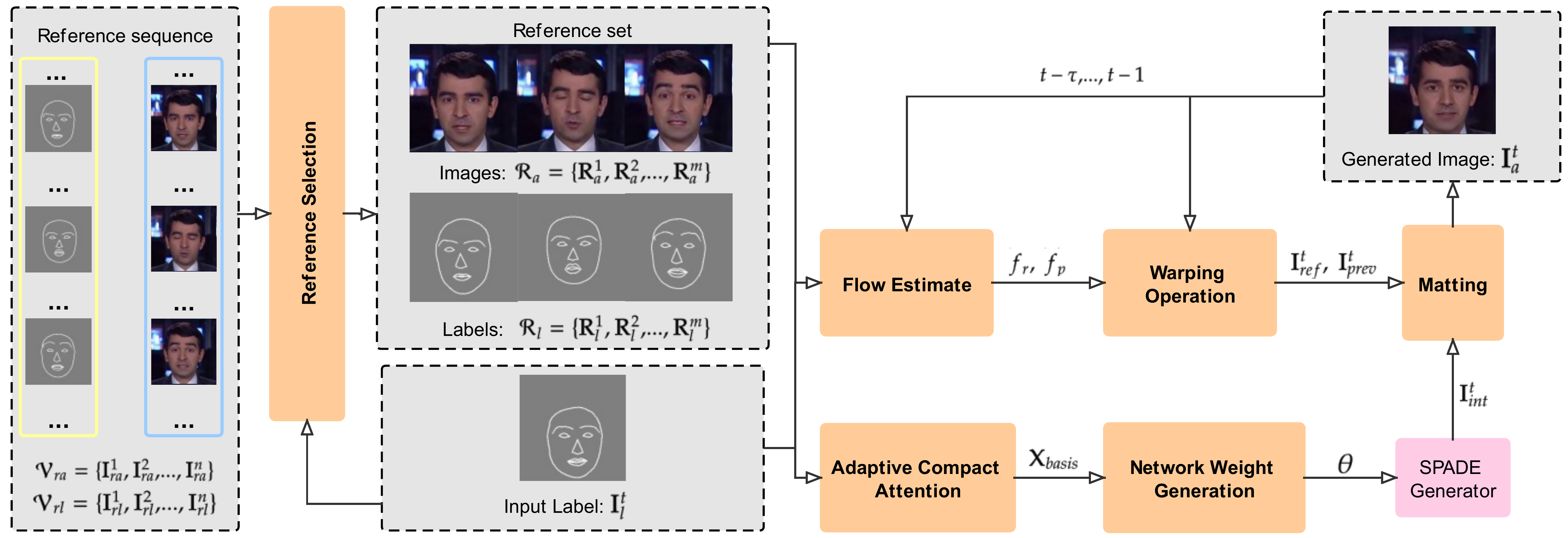}
    \caption{The pipeline of our method. At time step $t$, the model takes as input an input label $\mathbf{I}_l^t$, a reference appearance image set $\mathcal{R}_a = \{\mathbf{R}_a^1, \mathbf{R}_a^2, ..., \mathbf{R}_a^m\}$ and the corresponding reference semantic images set $\mathcal{R}_l = \{\mathbf{R}_l^1, \mathbf{R}_l^2, ..., \mathbf{R}_l^m\}$. The model first conducts an adaptive compact attention on these inputs to obtain the contextualized reference image features $\mathbf{X}_{basis}$. The features are then leveraged to generate network weights $\theta$ for a SPADE generator. The generator further outputs an intermediate image $\mathbf{I}_{int}^t$ of the input label $\mathbf{I}_l^t$. In the meantime, we estimate the optical flow $f_r$ and $f_p$ of both the reference appearance images $\mathbf{R}_a$ and previous synthesized images $\{\mathbf{I}_a^{t-\tau}, ..., \mathbf{I}_a^{t-1}\}$. We utilize these optical flows to warp these images and output the warped images $\mathbf{I}_{ref}^t$ and $\mathbf{I}_{prev}^t$. Finally, we obtain a synthesized appearance image $\mathbf{I}_a^{t}$ by aggregating the warped images $\mathbf{I}_{ref}^t$ and $\mathbf{I}_{prev}^t$ with the intermediate images $\mathbf{I}_{int}^t$ through a matting function. In inference phase, we additionally apply a reference selection method on a large reference sequence to dynamically construct a resourceful reference set according to the input label.}
    \label{fig:pipeline}
\end{figure}

Therefore, we propose an adaptive compact attention model for few-shot video-to-video translation, which can efficiently extract contextual features jointly from multiple reference images. Our key idea is to extract compact basis sets from all the reference images as a global representation, which encodes the correlations in the reference images while significantly reduces the computational cost. This basis extraction process is shared between semantic and appearance reference images to further improve its efficiency. Moreover, by simultaneously considering all reference images, the representational power of the computed attention basis set is significantly enhanced and the redundancy of the extracted features are reduced. During testing, different from \cite{DBLP:conf/nips/Wang0TLCK19} that needs to manually select the reference images, we employ a novel reference selection method based on the Delaunay Triangulation algorithm to determine the resourceful references automatically. This selection scheme allows various appearance information encoded in the reference sequences to be fully exploited for more reliable video synthesis. The overview of our method is shown in Fig.~\ref{fig:pipeline}.

We validate our method on FaceForensics \cite{DBLP:conf/iccv/RosslerCVRTN19}, a large-scale talking-head video dataset and a human dancing video dataset collected from Bilibili\footnote{A large video sharing website in China. \url{https://www.bilibili.com}}. Extensive quantitative and qualitative evaluations demonstrate the superior performance and efficiency of the proposed method in producing photorealistic and temporally consistent videos. Comparisons to the related methods also show our remarkable improvements over the state-of-the-art method.

\section{Related work}\label{related-sec}
\textbf{Video generative model.}
Current methods for video generation can be roughly classified into three categories: unconditional video generation model, video prediction model, and video-to-video translation model. Unconditional video generation models \cite{uvideo1, uvideo2, uvideo3} focus on converting one or multiple random vectors to a video. Although different vectors can generate various videos, these methods lack explicit control over the generated videos. As for video prediction models \cite{video_predict_1, video_predict_2, video_predict_3, video_predict_4, video_predict_5, futureGAN, ContextVP}, a common pattern is to learn to predict future frames based on the current and previous frames. these methods also lack the flexibility to control the generation process since the future frames are simply forecasted by a learned prior knowledge. The most relevant methods to this work are video-to-video translation models \cite{DBLP:conf/nips/Wang0TLCK19, dancenow, vid2vid}, which aim to generate videos by converting semantic videos. Recently, Wang et al. \cite{DBLP:conf/nips/Wang0TLCK19} propose the first few-shot video-to-video translation method by adopting a dynamic weight generation scheme, which enables the network to deal with unseen domains. However, this scheme utilizes a pixel-wise attention to extract features from reference images; therefore it cannot capture higher-level context information in multiple reference images. Instead, our adaptive compact attention can efficiently extract global contextual features for producing more reliable results.

\textbf{Image-to-image translation.} 
Methods of this category aim to map an input image of a source domain to a target domain. The earliest image-to-image translation method can date back to the image analogies proposed by Hertzmann et al. \cite{analogies}. This method adopts a non-parametric way to translate an image to another style given a pair of examples. With the advances in deep learning, recent approaches rely on training parametric models such as CNNs \cite{colorization} or conditional GANs \cite{pix2pix} on a dataset of paired images to learn the translation function between two image domains. Image-to-image translation has been applied to various tasks, such as generating different poses of a human from different keypoints \cite{pose2image} or emotions from different faces \cite{ganimation}. Image-to-image translation can also be used to generate videos frame-by-frame, but they suffer from poor temporal-consistency since the generation of each frame is independent and there is no constraint between two adjacent frames.

\textbf{Attention mechanism.}
Attention mechanism is widely used in various fields such as machine translation, object detection, and semantic segmentation. Vaswani et al.\cite{attention} propose a self-attention module called transformer to extract the context feature at one position by aggregating features from all other positions in sentences.  Wang et al. \cite{non-local} propose a non-local neural network, which first introduces the self-attention mechanism to the field of computer vision. Li et al. \cite{attentive_objd} propose a global contextualized sub-network based on attention mechanism to extract the global contextual information in order to improve the performance of region-based object detectors. PSANet \cite{psanet} extracts contextual information from a predicted attention map. EMANet \cite{emanet} conducts the self-attention in an expectation-maximization manner by computing the basis for each pixel and combine the basis to obtain contextual features. Similarly, ACFNet \cite{acfnet} computes class centers for each pixel and aggregates each class center to form a class-level contextual representation. These methods aim to extract the intra-frame contextual information by self-attention. Instead, our goal is to explore higher-level inter-frame information, which requires to jointly consider the relations among reference semantic images, reference appearance reference, and input semantic images. Thus, directly applying these methods to this domain, i.e., few-shot vid2vid translation, is not feasible.

\section{Method}\label{method-sec}

In this paper, we focus on solving the problem of few-shot video-to-video translation. Given a $n$-frame input image sequence~$\mathcal{V}_l = \{\mathbf{I}_l^1, \mathbf{I}_l^2, ..., \mathbf{I}_l^n\}$ with semantic labels and a reference set, a photorealistic video $\mathcal{V}_a = \{\mathbf{I}_a^1, \mathbf{I}_a^2, ..., \mathbf{I}_a^n\}$ is synthesized with the same semantic labels as the input and the same appearance domain as the reference. In particular, $m$ appearance images $\mathcal{R}_a = \{\mathbf{R}_a^1, \mathbf{R}_a^2, ..., \mathbf{R}_a^m\}$ of target domain and their corresponding label images $\mathcal{R}_l = \{\mathbf{R}_l^1, \mathbf{R}_l^2, ..., \mathbf{R}_l^m\}$ constitute the reference set. By exploiting the attention mechanism, the existing few-shot video-to-video method \cite{DBLP:conf/nips/Wang0TLCK19} can generate the network weights for synthesizing a photorealistic video of previously unseen subjects depicted in the reference.

Nevertheless, the existing few-shot video-to-video translation method \cite{DBLP:conf/nips/Wang0TLCK19} does not consider the relations among multiple reference images, which leads to the loss of the motion-dependent and view-dependent information that will benefit the realistic video synthesis. Instead, we introduce a novel adaptive compact attention mechanism to extract contextual information from multiple reference images. Additionally, in order to exploit the various appearance information in such a sequence, we propose a reference selection method to automatically select a suitable reference set for each time step at the inference phase. In the remainder of this section, we describe the proposed method in detail. The overall pipeline is illustrated in Fig. \ref{fig:pipeline}.

\subsection{Adaptive compact attention}
\begin{figure}[t!]
    \centering
    \includegraphics[width = \linewidth]{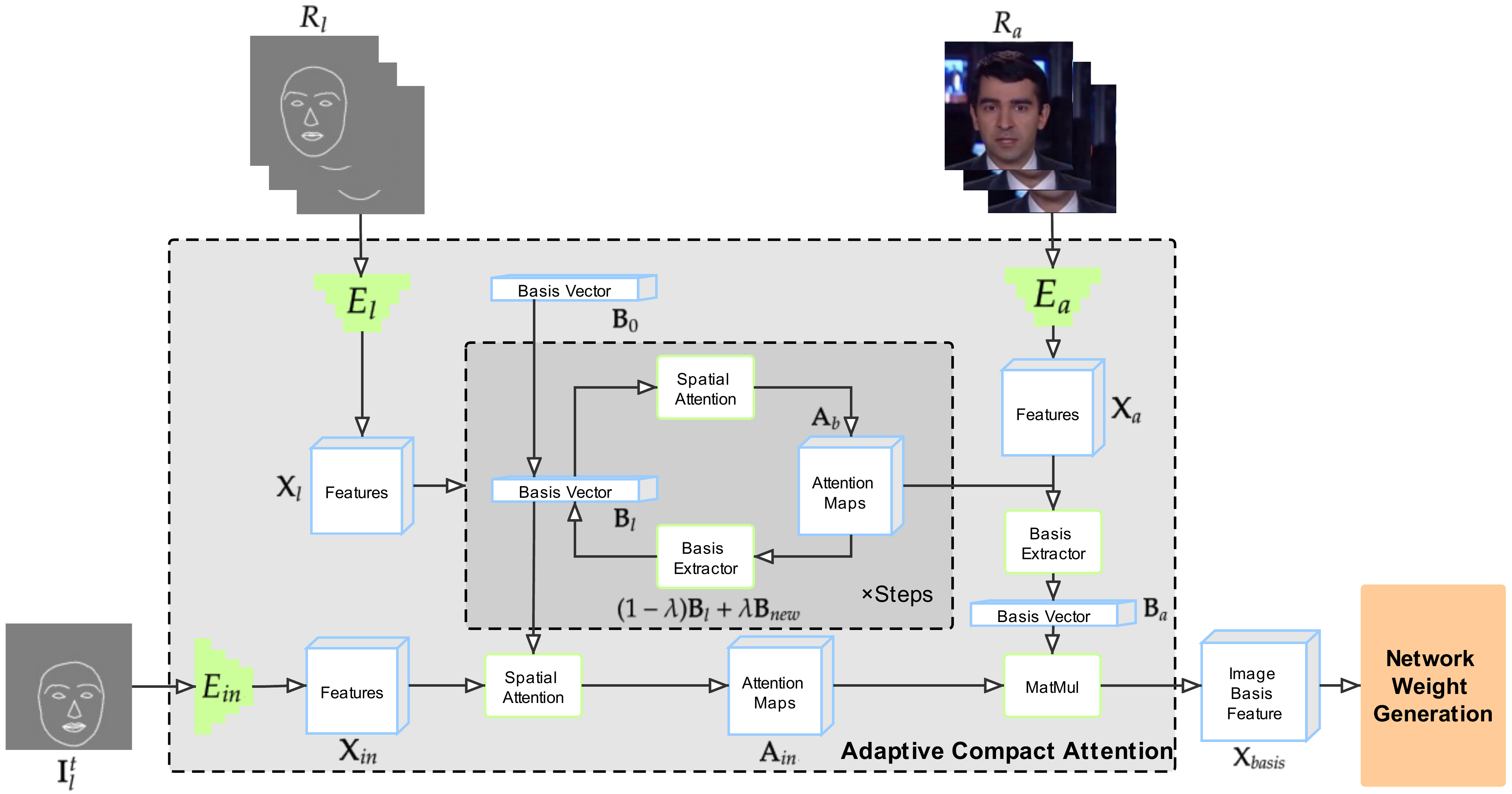}
    \caption{The adaptive compact attention takes an input label $\mathbf{I}_l^t$, a reference appearance image set $\mathcal{R}_a = \{\mathbf{R}_a^1, \mathbf{R}_a^2, ..., \mathbf{R}_a^m\}$ and corresponding reference semantic image set $\mathcal{R}_l = \{\mathbf{R}_l^1, \mathbf{R}_l^2, ..., \mathbf{R}_l^m\}$ as input and output contextualized image features $\mathbf{X}_{basis}$. It consists of three steps: (1) a feature extraction step to extract appearance/semantic features for each input by encoders with convolutional layers; (2) a basis extraction step that summarizes two basis sets $\mathbf{B}_l$ and $\mathbf{B}_a$ from reference appearance features $\mathbf{X}_a$ and reference semantic features $\mathbf{X}_l$; (3) a basis aggregation step combines these two basis sets with the features of input label $\mathbf{X}_{in}$ and outputs contextualized image features. This image basis features are further utilized for network weight generation.}
    \label{fig:model}
\end{figure}

To extract the contextual information from the reference set, we apply an adaptive compact attention mechanism rather than an image-by-image pixel-wise attention on the reference set for weight generation. As shown in Fig. \ref{fig:model}, our adaptive compact attention consists of three sequential steps: feature extraction, basis extraction, and basis aggregation. We now consider the moment $t$. To start with, the $\mathbf{I}_l^t$, $\mathcal{R}_a$ and $\mathcal{R}_l$ are fed to the encoders $E_a$, $E_l$ and $E_{in}$ to obtain the features. We then extract two basis sets of the features from $\mathcal{R}_a$ and $\mathcal{R}_l$ respectively. The basis sets are further aggregated with the features of input label $\mathbf{I}_l^t$ to form the output features that contain both the appearance and contextual information in the reference set.

\textbf{Feature extraction.} 
The feature extraction step is constructed by three sub-networks: a reference appearance encoder $E_a$, a reference label encoder $E_{l}$, and an input label encoder $E_{in}$. The encoders $E_a$ and $E_l$ take the reference appearance and label images as inputs and extract corresponding features $\mathbf{X}_a = [\mathbf{X}_a^1, \mathbf{X}_a^2, ..., \mathbf{X}_a^m]$ and $\mathbf{X}_l = [\mathbf{X}_l^1, \mathbf{X}_l^2, ..., \mathbf{X}_l^m]$, respectively. In the meantime, the encoder $E_{in}$ encodes the input image to feature $\mathbf{X}_{in}$. All encoders are equipped with $k$ convolutional layers and share the same architecture so that their outputs have the same shape, we utilize this fact to conveniently make these features interact with each other. In our case, $k$ is set to 5.

\textbf{Basis extraction.}
We operate basis extraction on features $\mathbf{X}_a$ and $\mathbf{X}_l$ yielded in feature extraction step. Suppose that each feature is of size $c \times h \times w$, where $h$, $w$ and $c$ represent height, width and channel numbers respectively. We have $\mathbf{X}_a \in \mathbb{R}^{m \times c \times h \times w}$, $\mathbf{X}_l \in \mathbb{R}^{m \times c \times h \times w}$ and $\mathbf{X}_{in} \in \mathbb{R}^{c \times h \times w}$.

We start by introducing the basis extraction on reference semantic features $\mathbf{X}_l$. First, we randomly generate an initialization $\mathbf{B}_0$ of the semantic bases $\mathbf{B}_l = [\mathbf{b}_1^T; \mathbf{b}_2^T; ...; \mathbf{b}_p^T], \mathbf{b}_i^T \in \mathbb{R}^{C}$, where $p$ is the number basis vectors, and reshape $\mathbf{X}_l$ to $hmw \times c$. Next, a spatial attention is applied between the reference semantic feature $\mathbf{X}_l$ and the semantic feature base $\mathbf{B}_l$, which yields an attention map $\mathbf{A}_{b} \in \mathbb{R}^{hmw \times p}$. Then, we obtain a new basis $\mathbf{B}_{new}$ by matrix multiplication between the attention map $\mathbf{A}_{b}$ and the reference semantic feature $\mathbf{X}_l$. Note we can apply a constraint such as normalization on the new basis to prevent significant change that may result in model collapse. We further update the semantic feature bases $\mathbf{B}_l$ with these new bases $\mathbf{B}_{new}$ based on a monument-based strategy. The spatial attention step and the basis update step are alternatively executed for $s$ times. In formula, we obtain the attention map $\mathbf{A}_{b}$ and semantic feature bases $\mathbf{B}_l$ as follows:
\begin{equation}
 \mathbf{A}_{b} = softmax(\mathbf{X}_l\mathbf{B}_l^T)
\end{equation}
\begin{equation}
 \mathbf{B}_{new} = norm(\mathbf{A}_{b}^T\mathbf{X}_l)
\end{equation}
\begin{equation}
 \mathbf{B}_l := (1-\lambda)\mathbf{B}_l + \lambda\mathbf{B}_{new}
\end{equation}
As for the basis extraction on reference appearance feature $\mathbf{X}_a$, instead of initializing bases and updating them iteratively, we borrow the previously calculated attention map $\mathbf{A}_{b}$, and directly acquire the appearance feature basis $\mathbf{B}_a$, as defined by
\begin{equation}
 \mathbf{B}_a = norm(\mathbf{A}_{b}^T\mathbf{X}_a)
\end{equation}

In summary, we extract the basis sets from the overall reference set to explicitly consider the relations among all reference images. These relations usually include motion-dependent and view-dependent information that are helpful to generate photorealistic videos. It is noteworthy that the number of bases $p$ here is far smaller than the pixel number of a feature map $m \times h \times w$. Moreover, as all the reference images are taken into account at the same time, we avoid extracting redundant information from different reference images. Furthermore, the complete process of the basis extraction only needs to be executed once as we share the attention map $\mathbf{A}_{b}$. In this way, the computational cost can be significantly reduced.

\textbf{Basis aggregation}
As aforementioned, our adaptive compact attention aims to extract a compact representation of reference images. We here obtain the desired compact representation by the basis aggregation. We conduct a spatial attention between semantic feature bases $\mathbf{B}_{l} \in \mathbb{R}^{p \times c}$ and reshaped input semantic features $\mathbf{X}_{in} \in \mathbb{R}^{hw \times c}$ to get an attention map $\mathbf{A}_{in} \in \mathbb{R}^{hw \times p}$. Then the attention map $\mathbf{A}_{in}$ is applied to the appearance features bases $\mathbf{B}_{a} \in \mathbb{R}^{p \times c}$ to obtain the final reference image features $\mathbf{X}_{basis} \in \mathbb{R}^{hw \times c}$. These features are reshaped to the same size of $c \times h \times w$ as the original input label features, which is defined by
\begin{equation}
 \mathbf{A}_{in} = softmax(\mathbf{X}_{in}\mathbf{B}_{l}^T)
\end{equation}
\begin{equation}
 \mathbf{X}_{basis} = \mathbf{A}_{in}\mathbf{B}_{a}
\end{equation}

Since the spatial attention is only conducted once between the low-rank $\mathbf{B}_{l}$ and $\mathbf{X}_{in}$ while \cite{DBLP:conf/nips/Wang0TLCK19} needs to calculate attention map for each reference image, our model has higher efficiency, even in cases where the reference set is large.

\subsection{Reference Selection}

\begin{figure}[t]
    \centering
    \includegraphics[width=\linewidth]{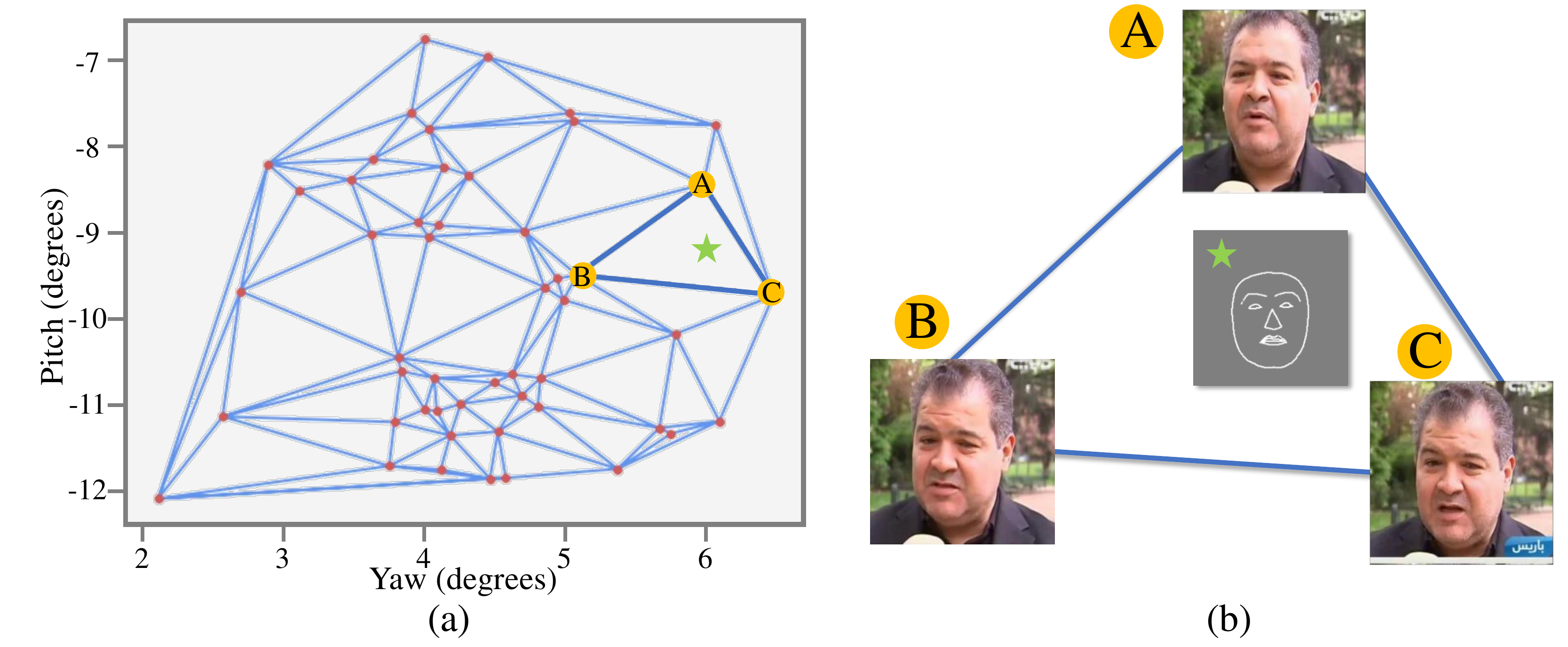}
    \caption{(a) An example of appearance map (mesh) of face reference sequence constructed by the Delaunay Triangulation algorithm. Each red point represents a reference image. (b) A reference selection result; given an input label, we can query in the appearance map to select resourceful reference images. }
    \label{fig:dau}
\end{figure}

To fully exploit the various information when a reference sequence is provided, we propose a reference selection method. Such a method can automatically select certain images from the reference sequence, which carries the representative information for video generation. Given a reference sequence with image sequence $\mathcal{V}_{ra} = \{\mathbf{I}_{ra}^1, \mathbf{I}_{ra}^2, ..., \mathbf{I}_{ra}^n\}$, and semantic image sequence $\mathcal{V}_{rl} = \{\mathbf{I}_{rl}^1, \mathbf{I}_{rl}^2, ..., \mathbf{I}_{rl}^n\}$, respectively, we construct an appearance map of the reference sequence, as shown in Fig. \ref{fig:dau}(a). Each image in the reference sequence is considered to be a point in this appearance map, with its coordinate $\mathbf{e}$ determined by feature extracted from the semantic image. For example, for each reference face, we can extract its Euler angle $(Pitch, Yaw, Roll)$  from the semantic image, and place it at $\mathbf{e} = (Pitch, Yaw)$ in a rectangular coordinate. Then we apply the Delaunay Triangulation algorithm to build a mesh structure on the reference sequence. 

At the inference phase, for an input semantic image $\mathbf{I}_l$, we query on the appearance map built on the reference sequence. First, we determine its coordinate $\mathbf{e}_l$ in the same way as aforementioned. Next, we search the appearance map to find the triangle $\mathcal{T}_l$ that contains $\mathbf{e}_l$. The vertices of $\mathcal{T}_l$ , $\{\mathbf{I}_{ra}^i, \mathbf{I}_{ra}^j \mathbf{I}_{ra}^k\}$ are selected as the reference set to join the synthesis process. One example of reference selection for an input face label is shown in Fig. \ref{fig:dau}(b). When more reference images are required, we find out the adjacent triangles that share a common edge with $\mathcal{T}_l$ and repeat the selection process. 
\section{Experiments}\label{experiments-sec}
\subsection{Datasets}
We train and evaluate our model on a large-scale talking-head video dataset FaceForensics \cite{DBLP:conf/iccv/RosslerCVRTN19} and a human dancing video dataset collected from Bilibili. The face dataset contains 1054 talking-head videos of different people. We utilize the open-source face detection library dlib \cite{dlib} to extract face landmarks as the semantic labels. The face dataset is split into 869 training videos and 185 testing videos. As for the human dancing video dataset, it consists of 1386 dancing videos from Bilibili. We apply the Openpose \cite{openpose} and Densepose \cite{densepose} to extract skeletons of people in dancing videos as the semantic labels. As these videos are posted by different users, the poses, background, and appearance of the person vary tremendously from one video to another, making it difficult to train on this dataset. For both the face video dataset and the human dancing video dataset, we omit frames if their semantic labels cannot be derived or are under poor condition, i.e., missing half of the key-points.

\subsection{Implementation details}
Our method adopts the same training strategy as the few-shot vid2vid method \cite{DBLP:conf/nips/Wang0TLCK19}. We train the models using ADAM optimizer \cite{adam} with $(\beta_1, \beta_2) = (0.5, 0.999)$. The learning rate is set to 0.0004 initially and linearly decayed to zero during the training. In the adaptive compact attention model, we set the number of basis $p$ to 128 and the corresponding number of iterations $s$ to 3 for the face dataset; since the synthesis of the dancing video requires more information, we increase the $p$ to 256 for the pose dataset. For each iteration in training, we randomly choose the input image and reference images in the same sequence to perform the translation. In the testing phase, we split out 20\% of each testing sequence as its reference sequence for reference selection. For the experiments without reference selection, the reference images are selected randomly from the reference sequence. It is notable that in the training of the pose dataset, we do not follow the few-shot vid2vid to add an additional face generator and discriminator because the finetuning of these two modules is rather challenging on this dataset.

\subsection{Baseline and evaluation metrics}

We compare our method against the state-of-the-art few-shot video-to-video translation approach~\cite{DBLP:conf/nips/Wang0TLCK19}. Furthermore, we employ an ablation study to demonstrate the effectiveness of different components of the proposed method. The metrics used for quantitative comparison in this paper include:

\textbf{FID (Fr\'echet Inception Distance)} \cite{fid} is calculated by computing the Fréchet distance between two Gaussian distributions fitted to feature representations of the Inception network. It measures the visual quality of generated images by calculating the similarity between generated images and real images.

\textbf{FVD (Fr\'echet Video Distance)} is similar to FID but is modified to apply on videos. It measures the quality of a generated sequence in both the visual and temporal domains.

\textbf{PSNR (Peak Signal to Noise Ratio)} is calculated between generated images and real images as a quality measurement. The higher the PSNR, the better the quality of the generated images is.

\textbf{Human Preference Score}. We conduct a user study where we generate 20 videos for each method and ask 15 people from different fields to choose their preferred videos.

\subsection{Results}

\begin{figure}[t]
    \centering
    \includegraphics[width=\linewidth]{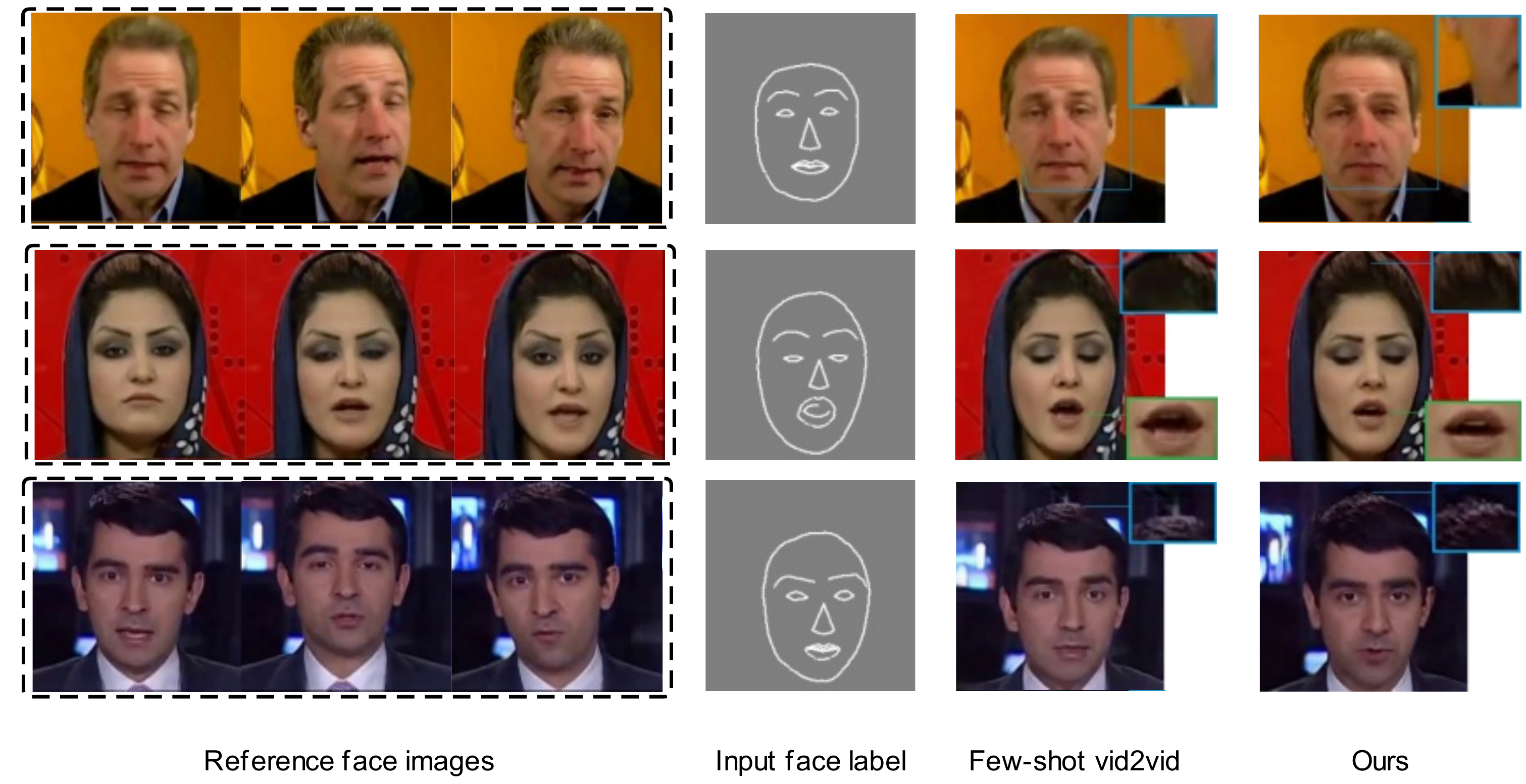}
    \caption{Comparisons of our method with few-shot vid2vid baseline for face video synthesis. Our method can generate sharper images compared to few-shot vid2vid, such as finer details in face boundary and mouth. }
    \label{fig:face_compare}
\end{figure}

\begin{figure}[t]
    \centering
    \includegraphics[width=\linewidth]{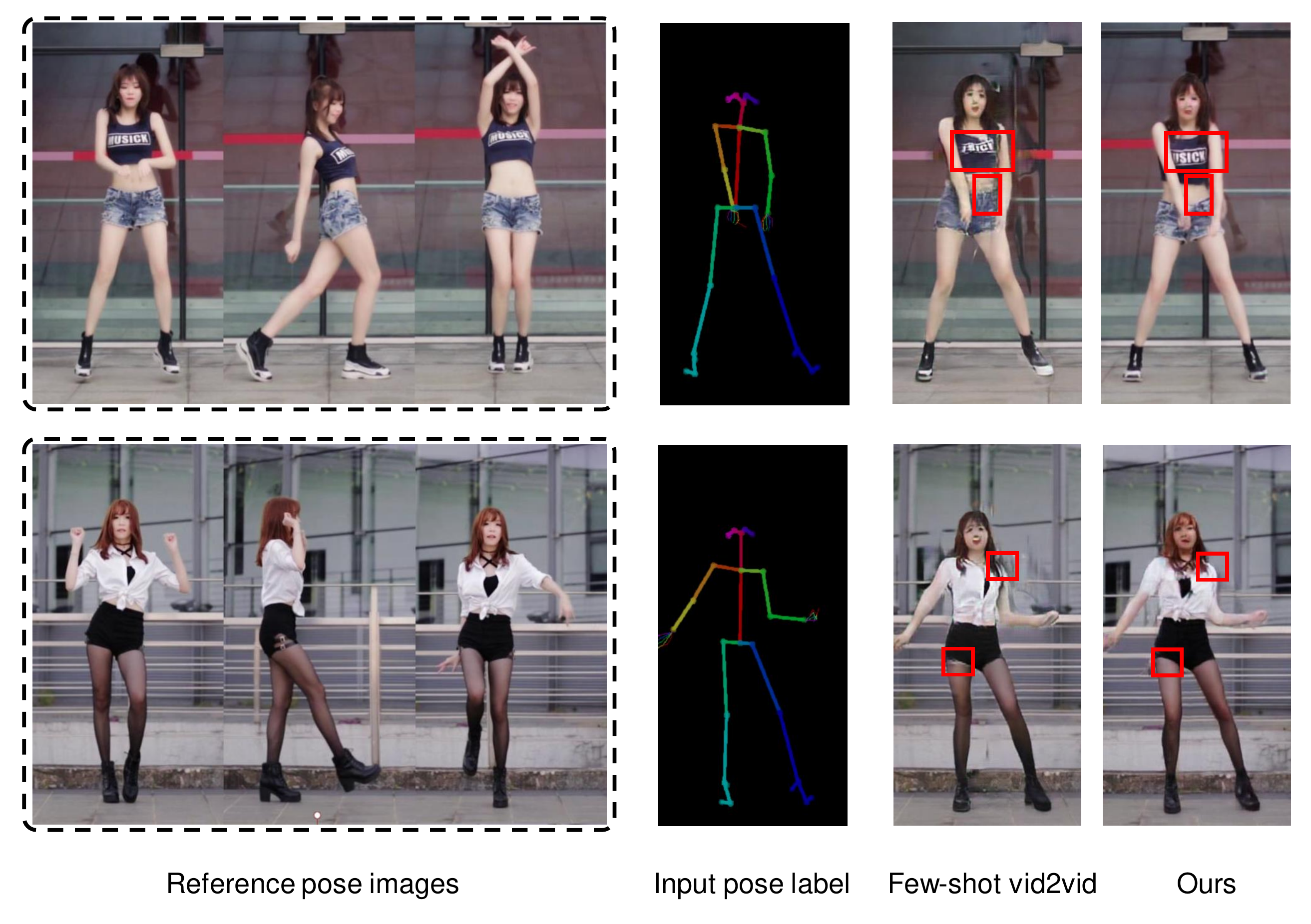}
    \caption{Comparisons of our method to few-shot vid2vid baseline for human video synthesis. It is seen clearly that our method can generate more correct details. }
    \label{fig:pose_compare}
\end{figure}

The qualitative comparison results with few-shot vid2vid \cite{DBLP:conf/nips/Wang0TLCK19} are shown in Fig. \ref{fig:face_compare} and \ref{fig:pose_compare}. It can be seen clearly that our model can generate videos with more details compared to the few-shot vid2vid method. For the face video experiment in Fig. \ref{fig:face_compare}, some artifacts are observed in the result of few-shot vid2vid, such as the blurry face boundary (row 1) and mouth (row 2), while our method clearly generates these details. As for the human video results in Fig. \ref{fig:pose_compare}, our model can synthesize human videos with correct poses while preserving more details than few-shot vid2vid, including details of clothes (Row 1: more texture on jeans; Row 2: a knot the hem of shirts) and hairs (our model generates hairs with a more similar color and style to the reference images).

The quantitative comparisons with the few-shot vid2vid \cite{DBLP:conf/nips/Wang0TLCK19} on the face dataset are shown in Table \ref{result-table}, in which they have identical reference sets randomly picked from a reference sequence. We observe considerable improvements of both the FID and PSNR metrics, indicating our method can generate videos with higher frame-level quality. Thus it demonstrates our adaptive compact attention can extract extra information such as view-dependent information that can help synthesize more realistic frames. Moreover, a lower FVD of our model shows that our model also outperforms the few-shot vid2vid in video-level quality, which proves that contextual information, such as motion-dependent information, can greatly benefit the temporal consistency of synthesized videos.
\begin{table}[ht!]
  \caption{We set up three settings in our experiment: (1) the few-shot vid2vid as baseline; (2) our model with the same reference sets as few-shot; (3) our model with the reference sets picked by the reference selection method. Our model shows consistent superior performance in all quantitive measures compared with the few-shot vid2vid method on the task of synthesizing face videos. }
  \label{result-table}
  \centering
  \begin{tabular}{rcccccccc}
    \toprule
    {} & \multicolumn{4}{c}{FaceForensics talking-head videos} \\
    \cmidrule(l){2-5} \cmidrule(l){6-9}
    Method           & FID            & FVD            & PSNR            & Human Pref. \\
    \hline
    Few-shot vid2vid & 72.90          & 16.34          & 17.92           & 0.20           \\
    Ours             & 64.40          & \textbf{15.36} & 18.88           & 0.33          \\
    Ours with RS     & \textbf{52.63} & 18.03          & \textbf{19.16}  & \textbf{0.47}           \\
    \bottomrule
  \end{tabular}
  \vspace{-0.2cm}
\end{table}

We conduct an ablation study to show the effectiveness of the proposed reference selection method. We compare the proposed reference selection strategy with the previous experiment where the reference set is chosen randomly. The results show significant improvements in both the FID and PSNR metrics, which means the selection method can correctly find a representative reference set for synthesizing each video frame. Note that such a selection procedure would frequently change the reference set. As a result, it makes the video less stable in the temporal dimension with a higher FVD. 

To evaluate the time efficiency, in Table \ref{time-table}, we show the mean training time of our model and few-shot vid2vid respectively for each iteration. It can be observed that our model takes less time for a single iteration, which verifies that our model achieves higher time efficiency compared to the few-shot vid2vid.

\begin{table}[ht!]
  \caption{Mean training time of each iteration on a single Tesla P40.}
  \label{time-table}
  \centering
  \begin{tabular}{rcc}
    \toprule
    {} & \multicolumn{2}{c}{Single iteration time(s)} \\
    \cmidrule(l){2-3}
    Methods     & FaceForensics  &  Dancing dataset     \\
    \midrule
    Few-shot vid2vid & 0.284   & 0.710\\
    Ours             & \textbf{0.240}   & \textbf{0.552}\\
    \bottomrule
  \end{tabular}
  \vspace{-0.2cm}
\end{table}
\section{Conclusion}\label{conclusion-sec}
We present an adaptive compact attention model for few-shot video-to-video translation. The adaptive compact attention is conducted on multiple reference images to extract contextual features, which is highly informative for synthesizing realistic videos in few-shot Vid2vid translation. The key idea is to extract a compact basis set as a global representation of the reference frames. In addition, we propose a reference selection method, which allows the various appearance information encoded in the reference sequences to be fully exploited for more reliable video synthesis. Quantitative and qualitative results show the superior performance and efficiency of our method compared to the state-of-the-art method.

\section*{Boarder Impact}
This research can benefit the development of automatic video synthesis technologies, which plays a vital role in a various range of industrial applications, such as synthesis or modification of video clips of specific objects to assist photography, generation of digital photorealistic faces or dance videos for augmented reality (AR) and virtual reality (VR) in communication and entertainment. For academic research, this research can be used to generate more data to support other computer vision tasks such as Image/Video forgery detection.

On the contrary, this research might be abused to generate fake videos that infringe on someone's privacy. To prevent this kind of issue from happening, we will be devoted to developing video forgery detection against our proposed method in the future.

\bibliographystyle{IEEEtran}
\bibliography{references}

\end{document}